\documentclass[12pt]{article}
\pdfoutput=1
\usepackage{color,amsmath,amssymb,exscale,psfrag,epsfig}
\usepackage{cite,color,url}
\usepackage{pdfpages}
\usepackage{graphicx}
\usepackage{slashed}
\usepackage{url}
\usepackage{makecell}
\usepackage[utf8]{inputenc}
\usepackage[T1]{fontenc}
\usepackage{xcolor}
\usepackage{accents}
\usepackage{centernot}
\usepackage{enumerate}
\usepackage{subfig}
\usepackage{float}
\usepackage{placeins}
\usepackage{floatpag}

\usepackage[colorlinks=true
,urlcolor=blue
,anchorcolor=blue
,citecolor=blue
,filecolor=blue
,linkcolor=blue
,menucolor=blue
,linktocpage=true
,pdfproducer=medialab
]{hyperref}
\input epsf
\usepackage{lineno}

\def\bea{\begin{eqnarray}}
\def\eea{\end{eqnarray}}

\definecolor{nicered}{rgb}{0.7,0.1,0.1}
\definecolor{nicegreen}{rgb}{0.1,0.5,0.1}
\hypersetup{colorlinks,citecolor= nicegreen,linkcolor= nicered}

\def\be{\begin{equation}}
\def\te{\end{equation}}
\def\ee{\end{equation}}
\def\ba{\begin{eqnarray}}
\def\bea{\begin{eqnarray}}

\def\tea{\end{eqnarray}}
\def\ea{\end{eqnarray}}
\def\eea{\end{eqnarray}}

\def\bfra{\begin{frame}}
\def\efra{\end{frame}}
\def\al#1\fal{\begin{align}#1\end{align}}
\def\bfra#1\efra{\begin{frame}#1\end{frame}}

\catcode`\@=11
\def\lsim{\mathrel{\mathpalette\@versim<}}
\def\gsim{\mathrel{\mathpalette\@versim>}}
\def\@versim#1#2{\vcenter{\offinterlineskip
\ialign{$\m@th#1\hfil##\hfil$\crcr#2\crcr\sim\crcr } }}
\catcode`\@=12
\catcode`@=12
\begin{document}
\thispagestyle{empty}
\begin{flushright}
ICAS 047/20
\end{flushright}
\vspace{0.1in}
\begin{center}
	{\Large \bf  Intelligent Arxiv:\\ Sort daily papers by learning users topics preference} \\
\vspace{0.2in}
{\bf Ezequiel Alvarez$^{(a)\dagger}$,
Federico Lamagna$^{(b,a)\ddag}$,
	Cesar Miquel$^{(c)\star}$
Manuel Szewc$^{(a)\diamond}$
}
\vspace{0.2in} \\
{\sl $^{(a)}$ International Center for Advanced Studies (ICAS) and CONICET, UNSAM,\\ 
	Campus Miguelete, 25 de Mayo y Francia, CP1650, San Martín, Buenos Aires, Argentina }
\\[1ex]
{\sl $^{(b)}$ Centro At\'omico Bariloche, Instituto Balseiro and CONICET\\
Av.\ Bustillo 9500, CP8400, S.\ C.\ de Bariloche, Argentina}
\\[1ex]
{\sl $^{(c)}$ Easytech, Parana 759, piso 3, CP1048, Buenos Aires, Aregntina}
\end{center}
\vspace{0.1in}

\begin{abstract}
	Current daily paper releases are becoming increasingly large and areas of research are growing in diversity. This makes it harder for scientists to keep up to date with current state of the art and identify relevant work within their lines of interest. The goal of this article is to address this problem using Machine Learning techniques.  We model a scientific paper to be built as a combination of different scientific knowledge from diverse topics into a new problem.  In light of this, we implement the unsupervised Machine Learning technique of Latent Dirichlet Allocation (LDA) on the corpus of papers in a given field to: {\it i)} define and extract underlying topics in the corpus; {\it ii)} get the topics weight vector for each paper in the corpus;  and {\it iii)} get the topics weight vector for new papers.  By registering papers preferred by a user, we build a user vector of weights using the information of the vectors of the selected papers.  Hence, by performing an inner product between the user vector and each paper in the daily Arxiv release, we can sort the papers according to the user preference on the underlying topics.  

	We have created the website {\tt IArxiv.org} where users can read sorted daily Arxiv releases (and more) while the algorithm learns each users preference, yielding a more accurate sorting every day.  Current {\tt IArxiv.org} version runs on Arxiv categories {\tt astro-ph, gr-qc, hep-ph} and {\tt hep-th} and we plan to extend to others.  We propose several new useful and relevant implementations to be additionally developed as well as new Machine Learning techniques beyond LDA to further improve the accuracy of this new tool.

\end{abstract}

\vspace*{2mm}
\noindent {\footnotesize E-mail:
{\tt 
$\dagger$ \href{mailto:sequi@unsam.edu.ar}{sequi@unsam.edu.ar},
$\ddag$ \href{mailto:federico.lamagna@cab.cnea.gov.ar}{federico.lamagna@cab.cnea.gov.ar},
$\star$ \href{mailto:miquel@easytech.com.ar}{miquel@easytech.com.ar},
$\diamond$ \href{mailto:mszewc@unsam.edu.ar}{mszewc@unsam.edu.ar}
}}

\newpage
\section{Introduction}
\label{section:1}
Machine Learning techniques are playing a crucial role in changing our every day experience with the World.  During the last years there has been a huge growth in the amount and in the class of applications of the different techniques and subfields within Machine Learning.  In particular, Natural Processing Language has been greatly developed by needs in the internet and communication industry.  The concept of {\it Topics Model} \cite{Blei:2012:PTM:2133806.2133826} has been extremely useful not only to understand and index a corpus of documents but also to learn compelling features about it.  This same technique has been adapted and used in other disciplines such as Bioinformatics \cite{bioinformatics}, Chemistry \cite{chemistry}, Health care \cite{health} and Physics \cite{Dillon:2019cqt,Metodiev:2018ftz,Alvarez:2019knh}, among others.  Along this work we pursue the objective of applying Topics Model to scientific literature and to use its outcome to create a new tool for sorting papers according to each user topics preference.

There exist a variety of algorithms applied to scientific papers and scientific literature.  Among them we can mention {\tt Arxitics.com} which allows to share voting, reviews and comments to provide an enhanced interface for reading and discussing the Arxiv; {\tt Scirate.com} which sorts papers according to ratings from the community, Ref.~\cite{mining} which mines scientific articles for recommending them to users based on abstract content using a personal collection of references, {\tt CiteULike} which allowed users to share preferences on scientific papers, {\tt Mendeley.com} which is a complete desktop service for sorting and archiving bibliography and generating bibliography for given articles, among other services, and Ref.~\cite{bleiwang} which uses collaborative filtering within a framework of topic modeling to recommend articles to users.   Some of these and other cases use rating algorithm, hand-made functions, and/or Machine Learning techniques to provide scientists with a better access to bibliography.  However, at current knowledge of the authors, there is not yet an available algorithm that learns from each user personal preferences and sorts the scientific papers accordingly for each user, which is the main goal that drives the content of this work.

Our departure point to tackle this problem is that in many cases the creation of a new scientific paper can be modeled as putting together scientific knowledge from different topics into a new problem.  This understanding of a scientific paper corresponds to the modeling of documents within a corpus in which each document is a specific mixture of topics in given proportions. The Latent Dirichlet Allocation (LDA) \cite{bleingjordan} algorithm is an unsupervised Machine Learning framework to address the topics and weights extraction in this kind of corpus.  On the other hand, and solely for the purposes of this article, we may also model the interests of scientists through these given topics:  scientists have a weight on each one of these topics that represents their interest on them. Given this modeling, it is suitable to start our enterprise of classifying papers and Arxiv readers based on the LDA algorithm.

This work is divided as follows.  In Section 2 we briefly describe the Topics Model and the Latent Dirichlet Allocation tool as an unsupervised Machine Learning algorithm to extract abstract topics from a given corpus of documents.  In Section 3 we apply Latent Dirichlet Allocation to a selected group of Arxiv categories and to each category itself.  We show how the study of these corpora through this tool provides a new mechanism to understand their constitution and a new tool for classifying them.  In Section 4 we present and discuss the website {\tt IArxiv.org} which allows Arxiv readers to access the bibliography sorted according to their topics preference.  In Section 5 we discuss the prospects for future implementations to {\tt IArxiv.org}.  Section 6 contains the conclusions of this work.

\section{Topics Model in Latent Dirichlet Allocation (LDA)}\label{section:2}

Topic Modeling, or Topics Model, is a general framework of statistical models which aims to infer abstract ``topics'' from a given corpus of unlabeled documents. These abstract topics can be thought of as the generators of the corpus and can thus be assumed to encode all the information necessary about the corpus. The topics can be used to label the documents with some criteria or to perform a dimensional reduction  from a large corpus to a relatively small number of topics over which to conduct different operations, in a similar way to other unsupervised clustering techniques such as Principal Component Analysis (PCA).

While in PCA the clustering works with correlations to find the principal variables that encode the variance of the data, topic modeling aims to find clusters of words with semantic meaning (although this is not guaranteed as we are using abstract topics). With these criteria in mind, we focus on probabilistic topic modeling where we assume a generative model that encodes the semantic structure of the corpus. These generative models can capture better inter- and intra-documents statistical structure than non-generative models such as Term Frequency times Inverse Document Frequency (TF.IDF) and are thus better suited for semantic clustering. From the generative model we can derive a posterior using Bayes theorem which, although intractable, can be approximated by algorithms such as Gibbs Sampling or Variational Bayes. The estimated posterior then provides the semantic information from which we can obtain the topic distributions over the vocabulary and each document distribution over the topics. One example of this, and the one we focus on this work, is the Latent Dirichlet Allocation (LDA) as detailed in Ref.~\cite{bleingjordan}.

In LDA we assume a mixed membership model with a fixed amount $K$ of topics, in which all of the $D$ documents are composed of every topic and every topic contains all the $N$ words in the vocabulary. To account for this, we assume each document $d$ has a multinomial probability distribution $\theta_d$ over the topic space and each topic $k$ has a multinomial distribution $\beta_{k}$ over the vocabulary. In turn, these probability distributions are sampled from two Dirichlet Distributions with hyperparameters $\alpha$ and $\eta$ respectively. In this context, $\theta$, $\beta$, $\alpha$ and $\eta$ play the role of latent variables which generate the corpus but are not directly observable. The Dirichlet distribution is the conjugate prior of the multinomial distribution, which allows the Bayesian inference to keep the multinomial distribution shape, albeit with different probabilities assigned to each category.

The procedure to generate the corpus is then the following:
\begin{enumerate}
\itemsep0em
\item For each topic $k=1,...,K$ sample a $\beta_k$ multinomial encoding the word proportions from the Dirichlet distribution $\text{Dir}(\eta)$.
\item For each document $d=1,...,D$, sample a $\theta_d$ multinomial encoding the topic proportions from the Dirichlet distribution $\text{Dir}(\alpha)$.
\item For each word $n=1,...,N$, sample a topic  $Z_{dn}$ from the multinomial distribution, $\theta_d$. 
\item Sample a word $w_{dn}$ from the topic distribution $\beta_{Z_{dn}}$.
\end{enumerate}
The corpus can then be simplified to two sets of distributions $\theta_{d}$ and $\beta_{k}$ sampled from two Dirichlet distributions, with each document generated by a set of two coin tosses per word, one to select a topic according to $\theta_{d}$ and one to select a word according to $\beta_{k}$  Further details can be found elsewhere \cite{Blei:2012:PTM:2133806.2133826}.

%

\begin{figure}[h!]
\centering
\includegraphics[width=0.75 \textwidth]{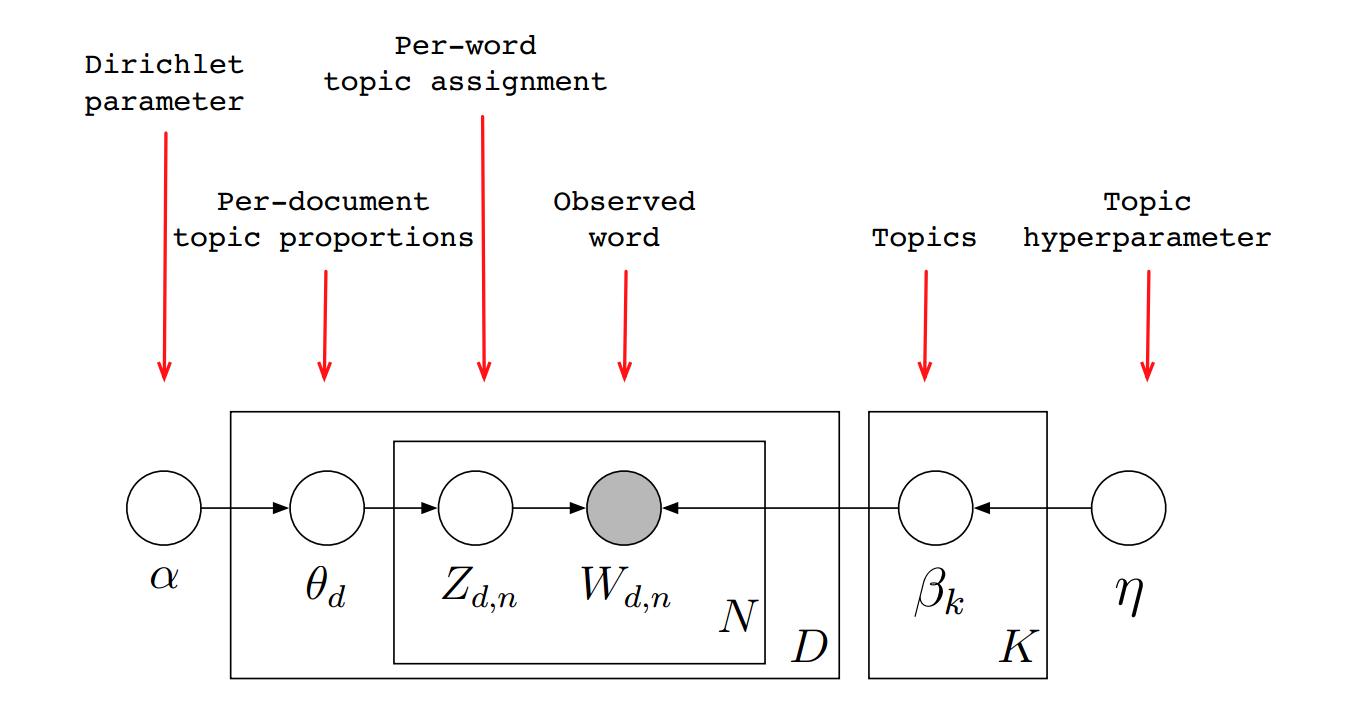}
\caption{Plate representation of the LDA procedure. As detailed in the text, a shaded circle is an observable variable while an unshaded circle is a latent variable. An enclosed plate denotes a repetition of the processes within. Figure obtained from Ref.~\cite{figplate}.}
\label{ldaplate}
\end{figure}

The procedure can also be encoded and understood in a plate diagram, where a circle represents variables of the model. If the circle is shaded, then the variable is observable while if it is unshaded, then the variable is latent and has to be inferred. The arrows indicate the direction of the sampling while the plates denote repetition over a number of steps indicated in the bottom right-hand corner. The plate diagram for LDA is shown in Fig.~\ref{ldaplate}.

While the generative model computes the probability of obtaining a word from a given set of variables $p(w,\theta,\beta,Z | \alpha, \eta )$, we are interested in the posterior probability $p(\theta, \beta, Z | w, \alpha, \eta )$ which allows to infer the topics and its distributions over the documents from the words. The posterior can be estimated by several means. We base our work on Online Variational Bayes as stated in Ref.~\cite{onlinelda} because it provides a relatively fast and efficient posterior estimation for large batches of documents.

All text processing and posterior estimation can be performed using the {\tt Gensim} package \cite{gensim,gensim2}, which implements the Online Variational Bayes method. We use this package  to obtain the probability distributions and the topic proportions in each paper. We do not infer $\alpha$ and $\eta$ but instead we keep them as priors from which we obtain the topic proportions in the documents $\theta$, the topic distributions over the vocabulary $\beta$ and the topic chosen to generate each word $Z$. As discussed in Ref.~\cite{onlinelda}, $\alpha$ and $\eta$ can also be updated. However, due to the large number of documents and topics, we are largely independent of the priors and we do not extract useful information from this update.

LDA performance has to be evaluated by taking into account convergence and validity of the obtained topics. While the former can be evaluated straightforwardly by measuring how the topics change with time, the latter is more arbitrary. The metrics we considered were perplexity and topic coherence. Perplexity is a measure of the models' ability to predict previously unseen data \cite{perplexity}, which can also account for LDA convergence as each new document should not produce a noticeable change in the topic structure once a good model is obtained. Topic coherence, on the other hand, is a measure of how the top words inside each topic share semantic similarity \cite{coherence} and can be used to evaluate the goodness of the models' semantic clustering.

\section{LDA on the Arxiv}

In this section we apply the LDA algorithm to the Arxiv database. We use the Bulk Data Access tool to download data from papers in categories {\tt astro-ph, gr-qc, hep-ph} and {\tt hep-th} of the last 8--10 years. We download paper id, title, abstract, submission date, authors, and the categories it belongs to. We use both title and abstract contents for the topics modeling. A starting point is to examine the Arxiv categories themselves, to study how well the LDA can find these categories as topics and how it performs as a classifier. Another, more ambitious task is to run an LDA model in each of these categories to find a topic structure in them, and analyze if it can help sort through the evergrowing number of articles in the scientific community. A large number of tasks can be performed with the use of this tool, for instance examine author affinity, or paper affinity, or even as a way of finding useful articles previously unknown to the user. Also it can help to study the evolution of certain themes within the community, or to see representative articles of a particular subject.

Before running LDA on the corpus of documents, a first processing has to be done on the text. This involves building the dictionary, and mapping the text into its bag-of-words form: this means that the order of the words in a text is no longer considered relevant, and texts become collections of words. The processing of the text for building the dictionary consists in two steps: word lemmatization and the removal of {\it stop words}. Word lemmatization, or stemming, refers to the process of keeping the stem of the words, and thus identifying words of the same origin. For instance, words ``running, ran, run, runs, runner'' become the same one ``run'', and thus preventing the proliferation of unique words with similar meaning in the dictionary \cite{lemm1,lemm2}. All punctuation and special characters are removed from the text, along with capitalization. The term stop words refers to those words that are the most common in natural language and are present in every text, like articles and prepositions. These words that carry syntactic (grammatical) information are not relevant on the topic structure of documents.  Techwords that would otherwise be removed or modified by the preprocessing but carry precise meaning are manually kept in the text.  A few examples of these are ``1-d'', ``2-d'', or particle names like ``e+'', or for instance ``AdS'', which would be modified to ``ad''. A list of these words is made by inspecting individual abstracts and the effect of the preprocessing on them. After this is done, an initial dictionary is created. However, the extremes of the word distribution are also removed. This implies words that are far too common, or extremely uncommon. We remove words that are present in less than 50 documents, and those that are present in more than 90\% of the corpus. After this is done, each document becomes a list of observed frequencies of words from the dictionary. The remaining dictionary size is around 2500 words.

\begin{figure}[h!]
  \begin{center}
    \subfloat[]{\includegraphics[width=0.4 \textwidth]{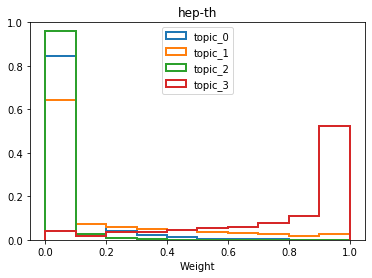}\label{hist4:hepth}}
    \hspace{3mm}
    \subfloat[]{\includegraphics[width=0.4 \textwidth]{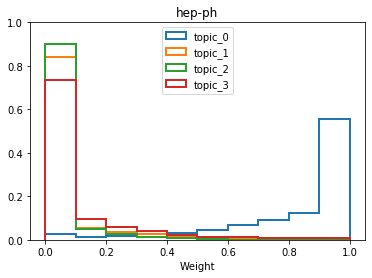}\label{hist4:hepph}}
    \\
    \subfloat[]{\includegraphics[width=0.4 \textwidth]{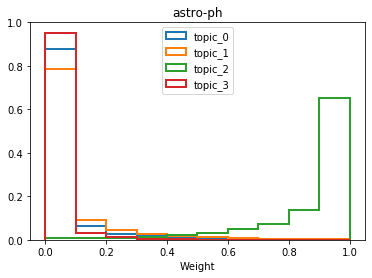}\label{hist4:astroph}}
    \hspace{3mm}
    \subfloat[]{\includegraphics[width=0.4 \textwidth]{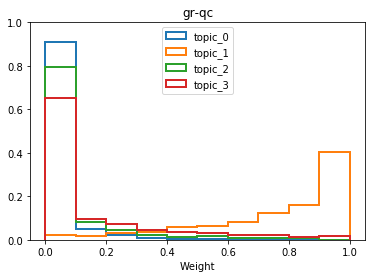}\label{hist4:grqc}}
    \\
  \end{center}
  \caption{Topic's weight distribution for papers in each category. A distinct topic arises over each one of the categories, indicating a match between unsupervised classification and categories definition.}
  \label{topic4}
  \end{figure}

A first exercise in the use of LDA is to check if it can classify between actual Arxiv categories. For this task we take samples of ``pure'' (that is, without cross-list) papers from each of the four categories {\tt astro-ph, gr-qc, hep-ph} and {\tt hep-th}, and we run an LDA model with four topics. For each category, we plot the topic proportions over its papers as a histogram, this can be seen in Fig. \ref{topic4}

\begin{figure}[h!]
  \begin{center}
    \subfloat[]{\includegraphics[width=0.4 \textwidth]{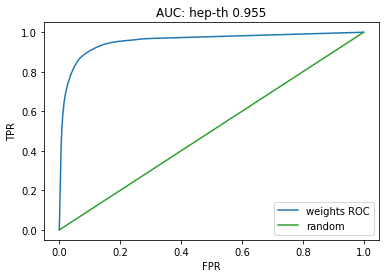}\label{roc:hepth}}
    \hspace{3mm}
    \subfloat[]{\includegraphics[width=0.4 \textwidth]{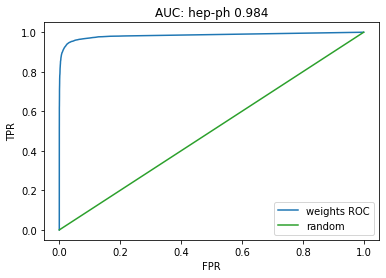}\label{roc:hepph}}
    \\
    \subfloat[]{\includegraphics[width=0.4 \textwidth]{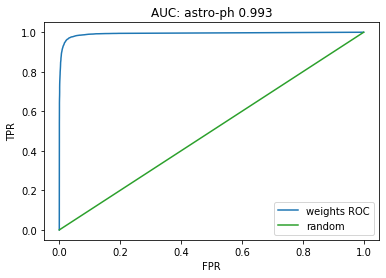}\label{roc:astroph}}
    \hspace{3mm}
    \subfloat[]{\includegraphics[width=0.4 \textwidth]{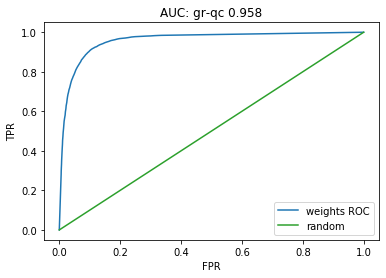}\label{roc:grqc}}
    \\
  \end{center}
  \caption{Receiver operating characteristic curves for a one-vs-all classification, for each of the four categories. Areas under curves are all higher than 0.95.}
  \label{roc4}
  \end{figure}

We see that for each category there is a topic that has a large proportion over the documents, and the rest of the topics have much smaller values. This means that these topic weights can be used for classification. To see how well this classifier can tag we plot the receiver operating characteristics (ROC) curves, one for each of the categories. The idea is that by using a threshold in the value of the weight, we can calculate the true positive rate and the false positive rate of one class against all others. We then plot the true positive rate against the false positive rate, for varying values of this threshold. A measure of the efficiency of the classifier is the area under the ROC curve, an area of 1 means an optimal classifier, whereas an area of 0.5 implies a random classifier. In Fig. (\ref{roc4}) we show these ROC curves for each of the four categories.
The areas under each of the curves show how good is the classifier and, in this case, all ROC curves have areas of around .95 and higher. This indicates that the {\it unsupervised} LDA algorithm has found topics that are in correspondence with the {\it defined} categories in the Arxiv database. This is an interesting result, as topics found by LDA are not required to have human interpretability. To examine topics further, we can analyze the top words occurring in each of the topics, as depicted in Table \ref{tabtopics}.

\begin{table}[h!]
\centering
\resizebox{\textwidth}{!}{%
\begin{tabular}{ | c | c |c | c |c | c |c | c |c | c |c | }
    \hline
    hep-th &  \thead{theori \\ 0.030} & \thead{field \\ 0.016} & \thead{gaug \\ 0.014} & \thead{gener\\ 0.009} & \thead{model\\0.009} & \thead{function\\0.009} & \thead{space\\0.009} & \thead{dimension\\0.009} & \thead{string\\0.008} & \thead{symmetri\\0.008} \\
    \hline
    hep-ph &  \thead{model \\ 0.021} & \thead{mass \\ 0.015} & \thead{neutrino \\ 0.011 }& \thead{decay \\ 0.010} & \thead{higg \\ 0.010} &\thead{quark \\ 0.009} & \thead{effect \\ 0.007} & \thead{product\\ 0.007} & \thead{dark\\ 0.007} & \thead{matter \\ 0.007} \\
    \hline
    astro-ph & \thead{star \\ 0.014} & \thead{observ \\ 0.013} & \thead{galaxi \\ 0.013} & \thead{mass \\ 0.010} & \thead{model \\ 0.008} & \thead{data \\ 0.006} & \thead{high \\ 0.006} & \thead{emiss \\ 0.006} & \thead{time \\ 0.006} & \thead{stellar \\ 0.005} \\
    \hline
    gr-qc & \thead{black \\ 0.019} &  \thead{ hole \\ 0.018} &  \thead{ field \\ 0.015} & \thead{model \\ 0.012} & \thead{equat \\ 0.011} & \thead{solut \\ 0.011} & \thead{ gravit \\ 0.011} &  \thead{graviti \\ 0.010} & \thead{energi \\ 0.010} & \thead{gener \\ 0.010} \\
    \hline
    \end{tabular}}
	\caption{Top ten most frequent words for each topic and their probabilities. First column indicates the Arxiv category that best matches each topic according to Fig.~\ref{topic4}.}
  \label{tabtopics}
\end{table}

We can see how the topics are different between themselves and how they can be interpreted. The fact that the algorithm was able to do this separation of topics in a completely unsupervised manner shows the strength of topic modeling.

To compare with more typical Machine Learning algorithms, we also study a multi layer perceptron for the task of classifying the documents. In this case, the learning is supervised, as the labels have to be used for the calculation of the loss function. We used documents in their bag of words form as input for the neural network.
That is, the input layer has a dimension equal to the size of the vocabulary, with each neuron corresponding to a different word, and its activation value corresponding to the number of times it appears in the document.
These neurons are then normalized, to work with values between 0 and 1.
We work with different architectures, changing the number and size of hidden layers.
The final layer has 4 neurons, equal to the number of classes. We used{ \tt categorical crossentropy} as the loss function, and a {\tt softmax} activation for this final layer.
We implement these architectures by using the Keras package \cite{chollet2015keras} for Python 3.
We find that already with a single hidden layer of 5 neurons the classification is as good as LDA, with ROCs having areas of about .95 and higher. With two hidden layers, of 5 and 50, the areas go upwards of .97 for all four topics.
We see that the unsupervised classifier using LDA has a performance as good as a supervised NN classifier.
For cases where tags are available there is usually no point in using unsupervised learning, however, inside each Arxiv category, there is a latent structure that is not made readily available by tags, and where an unsupervised Machine Learning algorithm like LDA is very useful.

\begin{figure}[h!]
  \begin{center}
    \subfloat[]{\includegraphics[width=0.3 \textwidth]{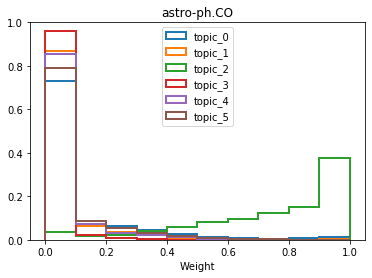}\label{hist:CO}}
    \hspace{3mm}
    \subfloat[]{\includegraphics[width=0.3 \textwidth]{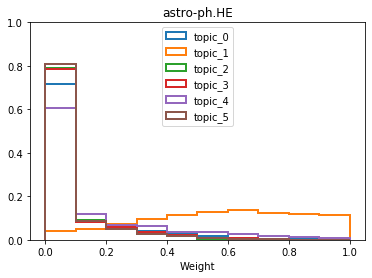}\label{hist:HE}}
    \hspace{3mm}
    \subfloat[]{\includegraphics[width=0.3 \textwidth]{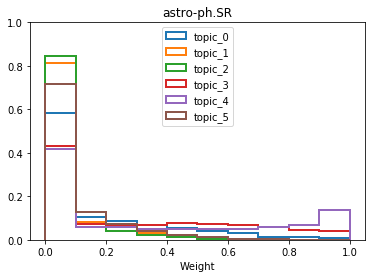}\label{hist:SR}}
    \\
    \subfloat[]{\includegraphics[width=0.3 \textwidth]{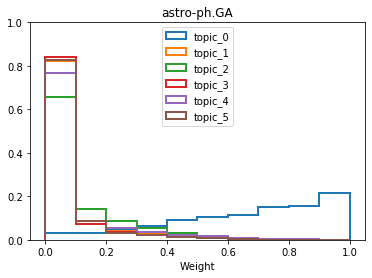}\label{hist:GA}}
    \hspace{3mm}
    \subfloat[]{\includegraphics[width=0.3 \textwidth]{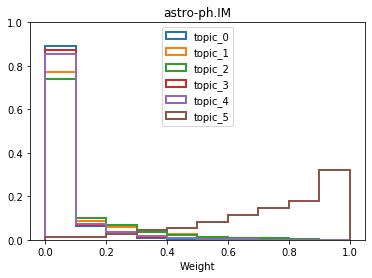}\label{hist:IM}}
    \hspace{3mm}
    \subfloat[]{\includegraphics[width=0.3 \textwidth]{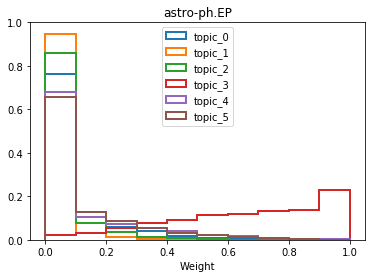}\label{hist:EP}}
    \\
  \end{center}
  \caption{Topic's weight distribution for papers in each subcategory of {\tt astro-ph} Arxiv category.}
  \label{astroph}
  \end{figure}

Taking this idea into account, and using the fact that some Arxiv categories do have subcategories available, we focus on {\tt astro-ph} and run an LDA model inside it. From 2009 onwards, this Arxiv category is split into 6 subcategories. We then take those papers, focusing once again on those without any cross-lists. We get a reasonably good clustering of the papers into 6 distinct topics, as can be seen in the histograms in Fig.\ref{astroph}
Once again as a measure of the classification capabilities of the model, we plot ROC curves for one-vs-all classification.
Except the {\tt astro-ph.SR} (Solar and Stellar) category, which has some non-negligible proportion of {\tt astro-ph.GA} (Galaxies) and {\tt astro-ph.EP} (Earth and Planetary) has an area under the ROC curve of about 0.7, the rest of the ROC curves have areas above 0.90.  One can visualize the topics by seeing the most frequent words in each topic in Table \ref{tabtopicsAP}.

\begin{table}[h!]
  \centering
  \resizebox{\textwidth}{!}{%
\begin{tabular}{ | c | c |c | c |c | c |c | c |c | c |c | }
  \hline
    CO & \thead{model \\ 0.018} & \thead{galaxi \\  0.016} & \thead{redshift \\  0.014} & \thead{cluster \\  0.012} & \thead{measur \\  0.012} & \thead{cosmolog \\  0.011} & \thead{scale \\  0.011} & \thead{dark \\  0.010} & \thead{paramet \\  0.010} & \thead{data \\  0.009} \\
 \hline 
 GA & \thead{star \\ 0.028} & \thead{galaxi \\  0.024} & \thead{mass \\  0.021} & \thead{ga \\  0.014} & \thead{stellar \\  0.012} & \thead{format \\  0.011} & \thead{line \\  0.010} & \thead{disk \\  0.009} & \thead{observ \\  0.009} & \thead{dust \\  0.009}  \\
 \hline 
 IM & \thead{data \\ 0.016} & \thead{telescop \\  0.015} & \thead{imag \\  0.015} & \thead{observ \\  0.010} & \thead{instrument \\  0.008} & \thead{high \\  0.008} & \thead{optic \\  0.007} & \thead{present \\  0.007} & \thead{base \\  0.007} & \thead{survey \\  0.007}  \\
 \hline 
 EP & \thead{planet \\ 0.027} & \thead{star \\  0.021} & \thead{orbit \\  0.020} & \thead{period \\  0.014} & \thead{mass \\  0.011} & \thead{observ \\  0.011} & \thead{binari \\  0.011} & \thead{model \\  0.008} & \thead{transit \\  0.008} & \thead{earth \\  0.007}  \\
 \hline 
 HE & \thead{ray \\ 0.023} & \thead{emiss \\  0.019} & \thead{observ \\  0.018} & \thead{energi \\  0.016} & \thead{sourc \\  0.016} & \thead{gamma \\  0.016} & \thead{x-ray \\  0.015} & \thead{radio \\  0.014} & \thead{time \\  0.011} & \thead{detect \\  0.010}  \\
 \hline 
 SR & \thead{magnet \\ 0.023} & \thead{field \\  0.019} & \thead{solar \\  0.013} & \thead{model \\  0.013} & \thead{observ \\  0.011} & \thead{simul \\  0.008} & \thead{region \\  0.008} & \thead{flare \\  0.007} & \thead{rotat \\  0.007} & \thead{structur \\  0.007} \\
 \hline 
    \end{tabular}}
	\caption{Top ten most frequent words for each topic for the astro-ph subcategories.  From top to bottom: Cosmology and Nongalactic Astrophysics (CO), Astrophysics of Galaxies (GA), Instrumentation and Methods for Astrophysics (IM), Earth and Planetary Astrophysics (EP), High Energy Astrophysical Phenomena (HE), and Solar and Stellar Astrophysics (SR).}
  \label{tabtopicsAP}
\end{table}

We conclude from this  exercise that the division into subcategories is slightly less sharp than the division of the whole Arxiv into categories, as the subcategories themselves might share more in common and have higher correlation between themselves.
In any case, as LDA is a mixed-membership model, it is designed to work better when the documents are composed of several topics simultaneously, otherwise for a single topic per document one could resort to simpler clustering algorithms. 
This property of LDA, along with the fact that some categories are not subdivided motivates us to use this topic modeling to explore the underlying structure of the papers within these categories.

Taking for example {\tt hep-ph}, we can run LDA with 40 topics, and then look at the topics themselves to see if they have interpretability.  
We select and show some of those topics in Table \ref{phtable}.
We can see how the model has captured distinct topics that are readily interpretable.
Here we included four different ones and label them according to the subjects they are about. 

\begin{table}[h!]
  \centering
  \resizebox{\textwidth}{!}{%
\begin{tabular}{ | c | c |c | c |c | c |c | c |c | c | c | }
  \hline
 lhc & \thead{lhc \\ 0.066} & \thead{search \\  0.043} & \thead{collid \\  0.042} & \thead{tev \\  0.033} & \thead{model \\  0.018} & \thead{energi \\  0.017} & \thead{decay \\  0.016} & \thead{mass \\  0.016} & \thead{signal \\  0.016} & \thead{hadron \\  0.015} \\ 
 \hline 
 higgs & \thead{higg \\ 0.136} & \thead{model \\  0.058} & \thead{boson \\  0.058} & \thead{coupl \\  0.033} & \thead{standard \\  0.028} & \thead{scalar \\  0.027} & \thead{electroweak \\  0.021} & \thead{sm \\  0.019} & \thead{mass \\  0.017} & \thead{h \\  0.017} \\ 
 \hline 
 lattice qcd & \thead{quark \\ 0.166} & \thead{qcd \\  0.064} & \thead{mass \\  0.055} & \thead{lattic \\  0.051} & \thead{gluon \\  0.029} & \thead{heavi \\  0.024} & \thead{light \\  0.023} & \thead{result \\  0.021} & \thead{hadron \\  0.019} & \thead{flavor \\  0.018} \\ 
 \hline 
 dark matter & \thead{dark \\ 0.156} & \thead{matter \\  0.148} & \thead{dm \\  0.026} & \thead{particl \\  0.024} & \thead{model \\  0.023} & \thead{interact \\  0.019} & \thead{detect \\  0.016} & \thead{annihil \\  0.015} & \thead{direct \\  0.015} & \thead{relic \\  0.014} \\ 
 \hline
    \end{tabular}}
      \caption{Top ten most frequent words for a few selected topics inside {\tt hep-ph}. Names in the first column correspond to our labeling of the topics.}
  \label{phtable}
\end{table}

As a remark on the LDA Topics Model power, it is interesting to notice that, for instance, among the 40 topics in {\tt hep-ph} there are four that have the word ``qcd'' among the 10 most likely ones.
Looking at the rest of the accompanying words, these four topics can be interpreted as corresponding to ``lattice qcd'', ``precision qcd'', ``qcd effective theories'' and ``qcd phase transition'', as it can be seen in  Table \ref{qcdtable}.  Similar conclusions are found with other keywords in this and in other categories.

\begin{table}[h!]
  \centering
  \resizebox{\textwidth}{!}{%
\begin{tabular}{ | c | c |c | c |c | c |c | c |c | c | c | }
  \hline
 lattice & \thead{ quark \\ 0.166} & \thead{ qcd \\   0.064} & \thead{ mass \\   0.055} & \thead{ lattic \\   0.051} & \thead{ gluon \\   0.029} & \thead{ heavi \\   0.024} & \thead{ light \\   0.023} & \thead{ result \\   0.021} & \thead{ hadron \\   0.019} & \thead{ flavor \\   0.018} \\ 
 \hline 
  precision & \thead{ jet \\  0.046} & \thead{ order \\   0.040} & \thead{ correct \\   0.039} & \thead{ lead \\   0.036} & \thead{ qcd \\   0.033} & \thead{ nlo \\   0.020} & \thead{ calcul \\   0.018} & \thead{ parton \\   0.017} & \thead{ soft \\   0.017} & \thead{ result \\   0.017} \\ 
 \hline 
  effective theory & \thead{ meson \\  0.101} & \thead{ vector \\   0.067} & \thead{ eta \\   0.061} & \thead{ rule \\   0.048} & \thead{ sum \\   0.043} & \thead{ rho \\   0.041} & \thead{ qcd \\   0.035} & \thead{ scalar \\   0.025} & \thead{ light \\   0.025} & \thead{ pseudoscalar \\   0.024} \\ 
 \hline 
  phase transition & \thead{ phase \\  0.063} & \thead{ temperatur \\   0.054} & \thead{ transit \\   0.038} & \thead{ model \\   0.029} & \thead{ potenti \\   0.026} & \thead{ finit \\   0.025} & \thead{ qcd \\   0.022} & \thead{ critic \\   0.021} & \thead{ chemic \\   0.019} & \thead{ thermal \\   0.014} \\ 
 \hline
\end{tabular}}
\caption{QCD topics present in {\tt hep-ph}}
\label{qcdtable}
\end{table}

We then conclude that Topics Model can help provide a way to sort through documents in a deeper way than simply looking at individual keywords.
As topics involve several words at the same time, the model can capture co-occurrence of keywords within topics, which is important for sorting papers according to personal preference in this space. 

\begin{figure}[h!]
\begin{center}
\subfloat[]{\includegraphics[width=0.4\textwidth]{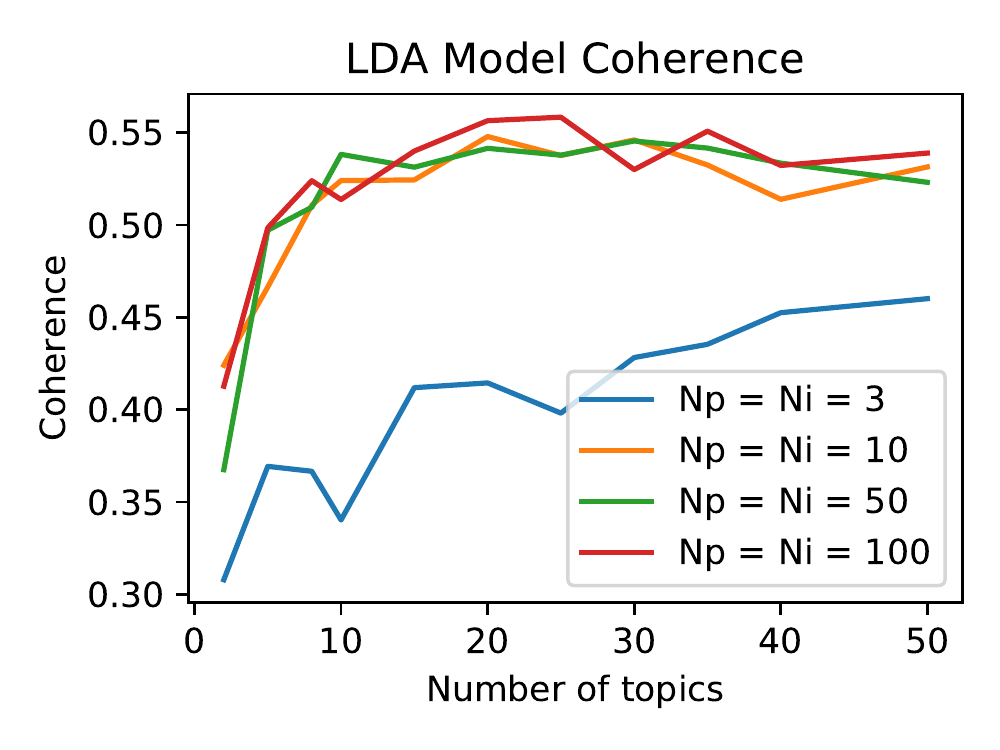}\label{metrics:coh}}\hspace{3mm}
\subfloat[]{\includegraphics[width=0.4\textwidth]{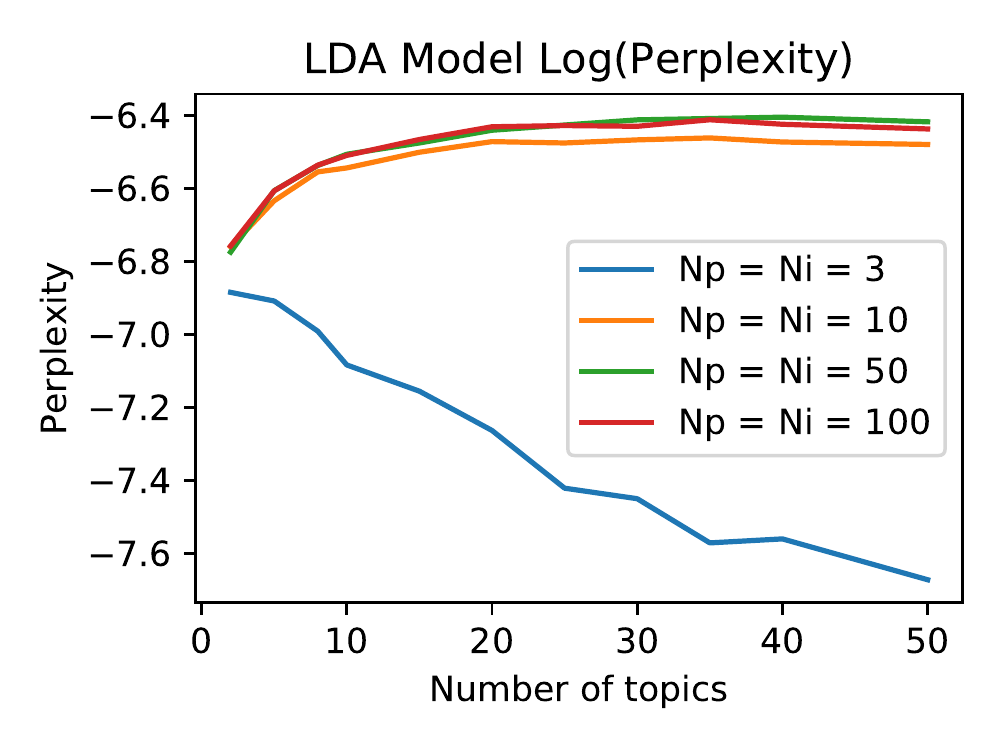}\label{metrics:perp}} \\
\caption{a) Coherence plot and b) Log of Perplexity. Each of the metrics is plotted against the number of topics, and for different values of passes and iterations..}
\label{metrics}
\end{center}
\end{figure}

As mentioned in section~\ref{section:2}, we consider different metrics to measure the goodness of the LDA Topics Model, in a sense allowing us to perform a scan over the hyperparameters and tune them to their optimal values.
In particular, we plot both perplexity and topic coherence in Fig.~\ref{metrics}.
Both plots have been computed using the {\tt CoherenceModel.get\_coherence()} and {\tt log\_perplexity()} functions from the {\tt Gensim} Package \cite{gensim2}, respectively.  As it can be seen from these results, specially from the perplexity plot, a good model convergence requires above $\sim 30$ topics and above $\sim 50$ passes and iterations.  In our LDA models we have used number of topics ranging from 30 to 60 and passes and iterations above 100.

\begin{figure}[h!]
  \begin{center}
    \subfloat[]{\includegraphics[width=0.4 \textwidth]{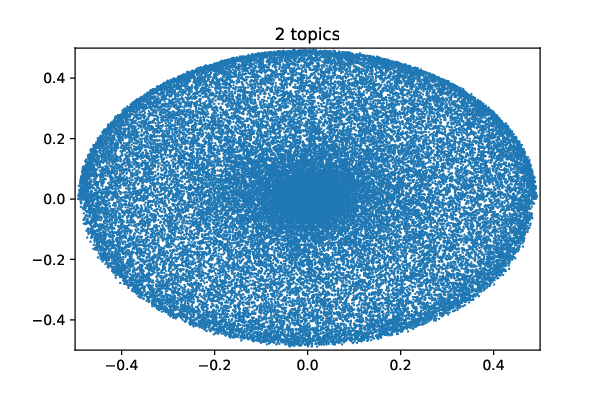}\label{pizza:2t}}
    \hspace{3mm}
    \subfloat[]{\includegraphics[width=0.4 \textwidth]{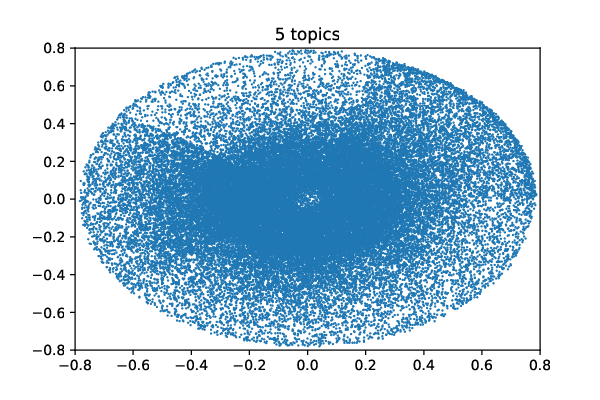}\label{pizza:5t}}
    \\
    \subfloat[]{\includegraphics[width=0.4 \textwidth]{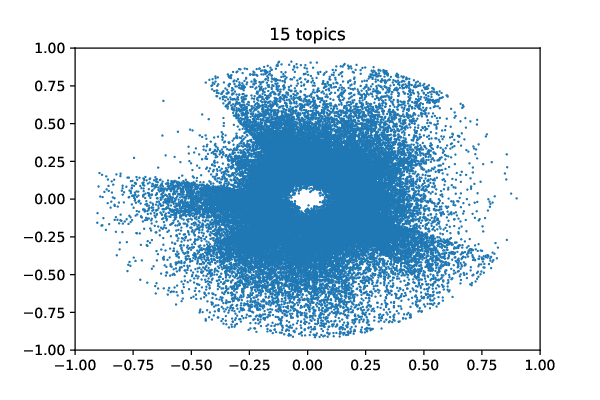}\label{pizza:15t}}
    \hspace{3mm}
    \subfloat[]{\includegraphics[width=0.4 \textwidth]{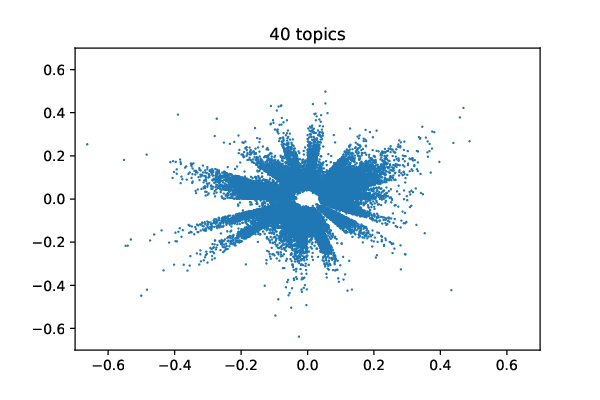}\label{pizza:40t}}
    \\
  \end{center}
	\caption{{\it Pizza-plots} (see text) for different values of $N_{t}$. We can see that as $N_t$ increases, documents get spread in a more asymmetric manner, and farther from the center.}
  \label{pizzas}
  \end{figure}

Another way to measure the model for different $N_{topics}$ values is by looking at what we coined as {\it pizza-plots}, which contain visual information on some of the model features.
These plots are defined as follows.
A circle centered in the origin is divided into slices and each topic is identified with one of these slices.
Each document in the corpus is a point that is located in the slice corresponding to its main topic.
The distance to the center of the circle for each point is defined as $ \sum_i \left( w^d_i - \frac{1}{N_t} \right)^2$, where $w^d_i$ is the weight of document $d$ in topic $i$.
This distance indicates the inhomogeneity of the document in the topics: the closer to the center is the point, the more homogeneous in the topics is the document.  For the sake of a better visualization, the angle of each point within each slice is random.
Fig.~\ref{pizzas} contains four pizza plots for $N_t=2,\ 4,\ 15$ and $40$ topics.
As it can be seen in the plots, as $N_t$ increases one obtains a clearer center (no documents uniformly spread in topics) and a clearer crust (no monotopic documents).
One can also see that as the number of topics increases, the documents get more asymmetrically distributed in the topics, as expected in these models.

Using the above results and conclusions, we have constructed four unsupervised LDA Topics Model with each one of the categories {\tt astro-ph, gr-qc, hep-ph} and {\tt hep-th} with number of topics ranging from 30 to 60 and using more than 100 passes and iterations in each construction.  These LDA Topics Models provide us with a vector of weights on each topic for each paper in the corpus and allow us to extract a new vector of weights for each new paper in each category.  These tools are the key for sorting papers for each user according to next Section discussion.



\section{IArxiv.org}

To allow users to navigate the corpus of papers from {\tt Arxiv.org} according to their preferences we built a website where they can register and view the listings from four of the Arxiv categories: {\tt astro-ph, gr-qc, hep-ph} and {\tt hep-th}. When a user starts browsing the papers we register each user preference by saving which papers the user is accessing. In this section we describe how the site works and what technologies we used to build it.

Once user have followed a standard registration process they can access their user preferences to indicate which categories they are interested in reading. The main page listing allows the user to filter the papers by date and displays a listing of title, abstract and authors. The papers are sorted according to the users preference as follows.   Each user has initially a vector (per Arxiv category the user is following) which has null components for each topic. As the user opens the different papers the site records which paper was opened and uses this information to update each user vector as explained below.  To sort a set of papers for a given user, the system computes the scalar product of the user vector with each one of the papers in the set and sorts them accordingly to their results, which is a set of real numbers.  Users can read sorted papers either through the daily Arxiv release, or can also choose specific slices of days.  The latter being very convenient for after a period of not reading the Arxiv.

The website is mobile friendly so that users can easily navigate it using their cellphones and the interface is minimal to reduce clutter and improve usability.

Current users vector computation is performed as follows.  We add the paper vector to the user vector for each paper in which the user is author, weighting as more important recent works.  We also add a paper vector to the user vector for each click that opens a paper and for each click that expands an abstract.  In these last operations we weight as more important opening the PDF than expanding an abstract.  The system is designed to give more weight to recent papers as time goes by. 

In terms of technology, the website is separated into a frontend web application developed using the ReactJS ({\tt https://reactjs.org/}) JavaScript framework and a backend server written in Golang ({\tt https://golang.org/}). All the information is stored on a PostgreSQL database that is accessed through the Golang backend. On a nightly basis a Python process reads all the new papers from the Arxiv and adds them into the database. The process computes the paper vectors.  
The database model is relatively straightforward: it has several tables to store users, papers, authors, paper vectors and user vectors. The information is normalized to reduce redundancy. Both backend and frontend are deployed using Docker ({\tt https://www.docker.com/}) into an EC2 instance in Amazon Web Services. This infrastructure will allow us to scale easily if we have an increase of traffic.

\section{Discussion and future prospects}

{\tt IArxiv.org} is currently running and providing sorted daily Arxiv papers according to each user preference, as described above.  However, having built this new tool using Machine Learning techniques over the Arxiv opens a new landscape of possibilities that could still improve the interaction between scientists and bibliography, and scientists among them, beyond the expected.

In a first phase for future upgrades we plan the following add-ons.  We will implement an optional daily IArxiv e-mail, in which users receive an e-mail with the papers sorted according to their preferences.  This e-mail should arrive as soon as the papers are available at Arxiv.  We also plan to include in the website a system that facilitates users to read many categories simultaneously meanwhile controlling the amount of papers selected in the release.  This will consist in an optional system of weighting and thresholds to be applied to each category.

Once the above goals are accomplished and implemented, we propose the following objectives, lines of research and work.

A direct and simple improvement to be performed at {\tt IArxiv.org} is to extend its scope to the remaining Arxiv categories.  This duty, however, requires a set of scientists in these categories to test its correct functioning during a reasonable period of time before opening it for the general public.

It is also planned to go beyond the LDA technique in the sorting of papers.  A compelling possibility is to construct an algorithm that recognizes authors affinity.  Bringing in this way a slight tunable increase to the score of papers in which  --for instance-- some of the authors have been previous co-authors of the IArxiv user.  Other tools to be studied and developed may also assist to recognize authors affinity.

There is also an interesting possible implementation in which {\tt IArxiv.org} can assist to find related bibliography to a given paper.  Within the LDA framework this is easily done by finding those papers in the corpus whose vectors have a large inner product with the desired paper.  We are currently testing this technique and considering implementing additional clustering techniques to provide a better outcome of this product.  Once this tool is consolidated, we will also implement an original variation consisting in the following: users can provide a new (not necessarily existing) title and abstract, and the system will return a set of papers in the corpus which are related by different mechanisms of clustering.

There are many other features which could be developed as byproducts of the above and/or from new tools.  As for instance, given two users who would be interested in starting a collaboration, the {\tt IArxiv.org} system could provide --as a guideline-- a set of papers in the corpus in which each paper contains simultaneously a large fraction of the topics preferred by the users.  Analogously, users could wonder in which affiliations they may find best affinity to their preferred topics.  Another potential interesting use is for seminars and workshops, where speakers may share their IArxiv topics vector and the scientists planning to attend may have a prior knowledge of their affinity to the speaker by requiring the scalar product of their vectors to the speaker vector through {\tt IArxiv.org}.

As part of {\tt IArxiv.org} maintenance we plan to rebuild the LDA models periodically in order to include new lines of research and experimental updates. In such cases we can rebuild all users vectors by re-applying their preferences within the new model topics.

\section{Conclusions}

We have studied the application of Topics Model to the corpus of Arxiv papers through the Latent Dirichlet Allocation unsupervised Machine Learning technique.  This has converted each paper into a topics weight vector which can be used to classify papers according to their distribution in topics.  As a byproduct of this study we have created the website {\tt IArxiv.org} which allows users to read daily Arxiv papers sorted according to their topics preference.

We have studied the Arxiv categories {\tt astro-ph, gr-qc, hep-ph} and {\tt hep-th} through their title and abstracts.  In a first analysis we have merged all categories into one corpus to run a LDA Model with four topics on it.  We have observed that this technique produces an excellent {\it unsupervised} separation of the four {\it defined} topics.  Motivated by this result, in a following step we have studied each category by separated.  We have analyzed variables as perplexity and coherence and determined that construction of good LDA models using {\tt Gensim} package would require number of topics ranging between $\sim$ 30--60, and no less than $\sim$50--100 passes and iterations.  We have also defined new plots to visualize the document distribution along the topics and verified that this number of topics provides a solid Topics Model for the studied corpora.

To create the website {\tt IArxiv.org} we provide each user with a new null vector of topics and we add to it weighted vectors of papers in which the user is author and papers that the user clicks to read in IArxiv web interface.  We have designed the system to provide more weight to more recent preferred papers.  This user vector contains therefore the user distribution of interest along the topics that describe the corpus.  To sort a daily Arxiv release of papers for a specific user, we compute the scalar product of the user vector with all the vectors of the papers in the release and sort them according to the outcome of this product.  

We have discussed future prospects for the {\tt IArxiv.org} system. In a next update we will provide the option of receiving a daily e-mail and introduce weight and threshold tools to facilitate users to read many categories simultaneously.  We also plan to create tools which would allow users to find relevant bibliography within the corpus of papers according to each specific topic preference.  We propose to study different Machine Learning techniques beyond LDA to further improve the accuracy of this tool, among other possible upgrades.

This novel system may find different applications in the scientific community.  As for instance scientific journals may find {\tt IArxiv.org} useful to classify and  distribute new papers to referees, if the latter have their user vector defined.

This new Machine Learning tool on the Arxiv provides an outstanding landscape of new tools and services for scientists.  We are open to new ideas and to scientists and institutions wishing to collaborate and/or partner in constructing a new bibliography engine for the benefit of everybody.

\section*{Acknowledgments}

We thank and congratulate {\tt Arxiv.org} for the outstanding duty of providing an excellent bibliography server and for sharing its content to Bulk Data Access through {\tt export.arxiv.org}.  We thank D.~de Florian for very valuable contributions and discussions on this project since it started.  We thank R.~Tito D'Agnolo, J.~Thaler, A.~Szynkman, E.~Roulet, J.~Kamenik, B.~Dillon and D.~Faroughy for fruitful conversations about the subject of this work.  We thank all colleagues who have used the {\tt IArxiv.org} website and provided crucial feedback for improving it.  We thank and congratulate A.~Cerqueiras for scripting IArxiv's front- and back-end. This project was conceived in Workshop Voyages Beyond the SM III.

\bibliographystyle{JHEP}
\bibliography{biblio}
\end{document}